\documentclass{article} 
\usepackage{iclr2026_conference,times}
\usepackage[utf8]{inputenc}
\usepackage[T1]{fontenc}
\usepackage{lmodern}
\usepackage{hyperref}
\usepackage{url}
\usepackage{graphicx}
\usepackage{longtable}
\usepackage{array}
\usepackage{booktabs}
\usepackage{pdflscape}
\usepackage{amssymb}
\usepackage{caption}
\usepackage{float}

\newcolumntype{L}[1]{>{\raggedright\arraybackslash}p{#1}}
\newcolumntype{C}[1]{>{\centering\arraybackslash}p{#1}}

\title{ShadowMiner v1 --- An Experience Report on Implementing and Measuring a Problem-and-Hypothesis Discovery Engine}

\author{Jinhyuk Choi \\
Inforience Inc. \\
Republic of Korea \\
\texttt{choi31u@inforience.com}}

\iclrfinalcopy
\begin{document}

\maketitle
\thispagestyle{fancy}
\lhead{}

\begin{abstract}
ShadowMiner v1 is a system that automatically discovers research problems and generates hypotheses from AI papers. It is a nine-stage pipeline. It structures documents into a knowledge graph and finds graph gaps in it --- structural blind spots in research. These graph gaps are included in the LLM generation prompt. Each generated hypothesis is then verified by checking whether it is already covered by existing research, scoring its quality, and checking that the facts it relies on are accurately drawn from its sources. This report does not propose a new generation or evaluation technique. It describes our experience of implementing and applying ideas from prior work, and measuring whether each one actually contributed.
\end{abstract}

\section{Introduction}

One of the first things a researcher does when starting new research is to read the related literature and work out ``which questions in this field have not yet been properly addressed.'' With thousands of papers appearing every month, carrying out this search entirely by hand is becoming ever harder. Recent years have seen several attempts to automate it with large language models (LLMs) (Section~\ref{sec:related}), and we have implemented and operate one of them: ShadowMiner, a system that takes AI papers as input and automatically extracts research problems and generates hypotheses.

This report is an experience report on what ShadowMiner builds, how it works, and how useful it turns out to be in practice. \textbf{It does not propose a new generation or evaluation technique.} Instead, we selected among previously published ideas (multi-hypothesis generation, critic ensembles, graph-gap detection, and others; Section~\ref{sec:techniques}), implemented and operated them, and then measured what contributed and what did not.

We ask three questions.

\begin{enumerate}
\item Do the hypotheses ShadowMiner produces match research that someone actually carried out later? (Section~\ref{sec:eval})
\item Does the quality score ShadowMiner assigns to its own hypotheses predict whether they actually appear in later papers? (Sections~\ref{sec:eval} and~\ref{sec:negative})
\item Does each design element that ShadowMiner borrowed actually contribute to the results? (Section~\ref{sec:techniques} and Appendix~\ref{app:techniques}; Section~\ref{sec:negative})
\end{enumerate}

To answer these questions, we rolled ShadowMiner's clock back to a past date (the ``freeze date''), had it generate hypotheses from only the literature up to that date, and checked whether each hypothesis appears in papers actually published after it. This check is our evaluation axis (Section~\ref{sec:setup}). This is a small-scale, early-stage study, and we have not yet run evaluations with human evaluators. The judgment of whether a hypothesis matches a later paper is therefore left to an LLM judge, and we evaluate how reliable that judge is (Section~\ref{sec:reliability}).

\paragraph{Organization.} Section~\ref{sec:related} reviews related work and Section~\ref{sec:system} describes the structure of ShadowMiner. Section~\ref{sec:techniques} covers the techniques ShadowMiner took from prior work, how they were implemented, and what results were obtained. Section~\ref{sec:eval} presents the evaluation results, and Section~\ref{sec:negative} covers improvement proposals that we tried but decided not to use. Section~\ref{sec:discussion} discusses limitations.

\section{Background and Related Work}
\label{sec:related}

\paragraph{Survey scope.} We reviewed more than 40 papers in a scholarly literature index that deal with ``generating research ideas and hypotheses with large language models.'' Of these, we examined in detail eight individual systems published in 2026 (EvoNarrator [12], VERITAS [38], AutoSOTA [5], EvoSci [13], DN-Hypo-Pipeline [8], GoR [15], Agentic-Ideation [1], EvoGens [11]); four fully autonomous systems that include execution and feedback (The AI Scientist [3], Execution-Grounded [9], ResearchEVO [31], Faraday [14]); four works focused on diversity (Barriers to Diversity [6], Anchorless Diversification [4], EvoDiverse [10], IDEAgent [21]); and four representative RL and multi-agent works (Graph-Native RL [16], SciDisco [33]/MPDS [27], HypoForge [19]). The earlier hypothesis-generation papers of 2023--2025 (ResearchAgent [30], SciMON [35], CoI-Agent [7], MOOSE-Chem [26], KG-CoI [24], SciAgents [32], AI Co-Scientist [2], HypoGeniC [20], HtT [18], Nova [28], Scideator [34]) are included for reference.

\paragraph{Comparison axes.} To place other systems on the same footing, we used eight axes: (1) hypothesis generation (producing and selecting candidates); (2) experiment execution (designing verification methods and actually running them); (3) feedback incorporation (a loop that feeds execution results back into generation); (4) autonomy (Tool $\rightarrow$ Analyst $\rightarrow$ Scientist, i.e., the degree of human involvement); (5) learning (prompts only / Supervised Fine-Tuning (SFT) / Reinforcement Learning (RL) / skill accumulation without fine-tuning); (6) diversity mechanisms; (7) evaluation method (single subjective rating / multiple judges / relative comparison / learned judge model); and (8) verifiability (whether a third party has a mechanism to verify the results).

\paragraph{ShadowMiner's position.} Table~\ref{tab:positions} places the systems on six of the eight axes, in four columns (axes 1--3 are grouped into one column). It lists 16 of the 20 systems examined in detail, and the 11 earlier papers on hypothesis generation (2023--2025) are summarized in a single row for reference. The columns mean the following. \textbf{Generate/Execute/Feedback} (grouping axes 1--3) marks, in that order, whether the system has a stage that produces hypothesis (idea) candidates, a stage that actually designs and runs experiments to verify the hypotheses, and a feedback loop that feeds execution results back into generation (O = present, $\times$ = absent). \textbf{Autonomy} (axis 4) is how much humans are involved, from Tool (a human operates every step) through Analyst (the human only gives a direction and the rest is automatic) to Scientist (from idea to result without a human). \textbf{Learning} (axis 5) is whether the model itself is changed: whether the system uses only prompts, changes weights through SFT or RL, or accumulates only reusable ``skills'' without fine-tuning. \textbf{Diversity mechanism} (axis 6) is whether there is a separate mechanism for producing several different candidates and, if so, what kind. Because the evaluation methods (axis 7) and verifiability (axis 8) are described so differently from paper to paper, we do not include them in the table.

\begin{table}[tb]
\caption{Positions of the systems on axes 1--6 (four columns: generate/execute/feedback, autonomy, learning, diversity mechanism).}
\label{tab:positions}
\centering
\footnotesize
\begin{tabular}{L{2.8cm}L{2.2cm}L{1.6cm}L{2.0cm}L{3.0cm}}
\toprule
System & Generate / Execute / Feedback & Autonomy & Learning & Diversity mechanism \\
\midrule
\textbf{ShadowMiner} & O / $\times$ / $\times$ & Tool--Analyst & Prompts only & None (parallel 4-fold generation and deduplication only) \\
Earlier hypothesis-generation papers (2023--2025, 11 papers, for reference) & O / mostly $\times$ / $\times$ & Tool--Analyst & Prompts & Mostly none \\
EvoNarrator [12] & O / $\times$ / $\times$ & Analyst & Prompts (graph constraints at inference time) & Macro-patterns of the evolution graph \\
GoR [15] & O / $\times$ / $\times$ & Tool--Analyst & SFT (graph $\rightarrow$ ideas) & None \\
Agentic-Ideation [1] & O / $\times$ / (self-critique) & Analyst--Scientist & SFT (agent trajectories) & None \\
EvoGens [11] & O / $\times$ / own scorer & Analyst & Prompts & Evolution (mutation + crossover) \\
EvoDiverse [10] & O / $\times$ / oracle & Analyst & Prompts & Two-temperature pool + exchange \\
IDEAgent [21] & O / $\times$ / critic loop & Analyst & Prompts & QD (Quality-Diversity) archive + lineage \\
AutoSOTA [5] $\cdot$ VERITAS [38] & O / O / O & Scientist / Analyst--Scientist & Prompts & None \\
The AI Scientist [3] & O / O / O & Scientist & Prompts & Idea archive (LLM mutation) + novelty filter \\
Execution-Grounded [9] & O / O / O & Analyst--Scientist & Prompts + RL/evolution & Evolutionary search \\
ResearchEVO [31] & Weak generation / O / O & Scientist & Prompts & Bidirectional co-evolution \\
Faraday [14] & O / O / O & Scientist & RL (rubric-judge reward) & None \\
Graph-Native RL [16] & O / $\times$ / $\times$ & Analyst & RL (GRPO) & Graph recombination \\
SciDisco [33] & O / O (data analysis) / O (turn-level) & Analyst--Scientist & RL (turn-level) & None \\
MPDS [27] & O / $\times$ / $\times$ & Analyst & Prompts & Persona debate \\
HypoForge [19] & O / O / O & Analyst--Scientist & Skill accumulation (no fine-tuning) & Generator--discriminator \\
\bottomrule
\end{tabular}
\end{table}

ShadowMiner starts from a training-free, prompt-only baseline configuration: in the first column it has \textbf{generation only (execution and feedback are $\times$)}, in the learning column it uses \textbf{prompts only}, and it has \textbf{no dedicated mechanism for inducing diversity yet}. Most 2026 systems add at least one of these. Here we group them by how they obtain performance, into three broad approaches: training-free, prompt-only methods (ShadowMiner also belongs here), learning-based methods that change the model itself, and evolution- and population-based methods that keep multiple candidates alive. Execution and feedback (the first column) are a separate axis and are not covered by this grouping.

\paragraph{Training-free, prompt-only methods} (including ShadowMiner, EvoNarrator [12], MPDS [27], and others) --- performance comes from prompts and pipeline design alone, without touching the LLM itself (its weights). This is light to implement and easy to reverse, but because the model does not change, performance stays tied to the base model. In Execution-Grounded [9], a search run without training the model stopped improving after the early rounds with two of the three models tested. How much these methods depend on the prompt still differs from system to system. Some designs give the LLM the background and the titles and abstracts of candidate papers and leave all of inspiration retrieval, hypothesis composition, and ranking to it (MOOSE-Chem [26]); ShadowMiner instead prepares the inputs itself before calling the LLM --- structuring documents into a knowledge graph and finding graph gaps --- and uses the LLM to draw out problems and hypotheses only in Stage 6.

\paragraph{Learning-based} (SFT in GoR [15] and Agentic-Ideation [1]; RL in Faraday [14], SciDisco [33], and Graph-Native RL [16]; skill accumulation in HypoForge [19]) --- the model itself is changed. Performance rises substantially in narrow domains, but ground-truth data for SFT is hard to obtain, and RL carries a risk of ``mode collapse'': as learning skews toward whatever raises the reward, the once-diverse outputs narrow to a few winning patterns. Execution-Grounded [9] observed exactly this in RL for idea generation: most sampled ideas converged on a few simple ones.

\paragraph{Evolution- and population-based} (EvoGens [11], EvoDiverse [10], IDEAgent [21], ResearchEVO [31]) --- these keep multiple candidates alive and raise diversity and quality together through mutation, crossover, and selection.

\paragraph{What we actually adopted.} ShadowMiner remains training-free (prompt-only) but took some low-cost elements from previous work. Parallel generation of multiple candidates (with a diversity metric) and multi-judge scoring are now standard in the pipeline; attaching sources to graph gaps was tested but not adopted (Section~\ref{sec:negative}). What was taken from where, how it was implemented, and how the results turned out are itemized in Table A1 (Appendix~\ref{app:techniques}).

\section{System}
\label{sec:system}

\subsection{Overview}

\paragraph{Definition.} Given a set of AI papers collected on a research topic, ShadowMiner automatically extracts the research problems that the field has not yet properly addressed and hypotheses that could plausibly answer them, and attaches supporting citations, a novelty verdict, and quality scores to each hypothesis. In effect, ShadowMiner automates the first stage of what a human researcher does: finding what has not been done yet and proposing what to try. The key point is that the output is grounded in the literature: the problems and hypotheses ShadowMiner produces are built on a graph of facts extracted from the input documents, and the facts each one relies on are marked with citation numbers.

\paragraph{Input and output.} The input is (1) the topic or query to analyze (e.g., ``improving the mathematical reasoning ability of large language models'') and (2) the set of AI papers that our own collection and retrieval system has gathered on that topic. The output is a number of problem statements (unresolved points) and, for each problem, candidate hypotheses (each consisting of a prediction, an expected result, and a verification method). Each hypothesis carries supporting citations \texttt{[n]}, a novelty verdict (exists/new/unclear), and four-axis quality scores (soundness, verifiability, groundedness, topical relevance).

\paragraph{Two design features.}

\paragraph{Generating and verifying hypotheses.} The input is prepared with care through document re-ranking, knowledge-graph extraction, normalization, and graph-gap detection and description (Stages 1--5). The LLM is then used to draw out problems and hypotheses in only one stage (Stage 6), which by default calls the LLM four times in parallel with the same prompt and merges the results (Section~\ref{sec:techniques}). Each generated hypothesis is then followed by three separate steps: novelty search, multi-judge scoring, and a citation-faithfulness check. Multi-judge scoring and the citation-faithfulness check call an LLM, whereas novelty search is embedding-based retrieval. Each of these replaces work that a human would otherwise do --- comparing against the literature to judge novelty, scoring quality, and verifying citations. At present, these steps are performed by an LLM or by embedding-based retrieval (Section~\ref{sec:rationale}).

\paragraph{Everything passes through the knowledge graph.} What goes into the generation prompt is not the original text of the papers but facts extracted from them and organized as a graph. In this graph, concepts that different papers express in different wording are merged into one (Stage 3, normalization), so facts from several papers can be attached to the same concept.

\paragraph{What ShadowMiner does not do.} It does not include an experiment-execution loop that actually runs a hypothesis; it addresses only hypothesis generation and evaluation.

\subsection{Knowledge Graph (KG)}

\paragraph{Why a graph.} The original text of a paper is not suitable for direct analysis across papers. ShadowMiner therefore extracts a knowledge graph from the text, a set of triples, and uses it instead. Each triple has the form ``concept A --- relation $\rightarrow$ concept B'' (e.g., ``Chain-of-Thought --- evaluated-on $\rightarrow$ GSM8K''), and it also records which sentence the triple came from. Once such triples accumulate, questions such as ``which datasets has Chain-of-Thought been evaluated on?'' can be answered directly, and this structure makes it possible to find graph gaps (Section~\ref{sec:pipeline}).

\paragraph{Schema.} For each paper we extract using a schema of six entity types (Task, Method, Dataset, Metric, Finding, Problem) and eight relation types (addresses, evaluated-on, improves, outperforms, uses, limited-by, addresses-problem, contradicts). The four entity types, Task, Method, Dataset, and Metric, together with the skeleton of relations, are borrowed from scientific-text information-extraction datasets (SciERC [43], SciREX [42], TDMSci [41], SciER [44]) and from the Papers With Code schema. The two entity types Finding and Problem are our own extension for hypothesis and gap analysis.

\paragraph{Narrow graph vs. wide graph.} From the stored triples we build two kinds of graph. The \textbf{narrow graph} contains only the triples that directly connect to the entities found in the input papers (one hop away), and it is included in the hypothesis-generation prompt. The \textbf{wide graph} extends to two hops and is used for graph-gap detection. For example, finding ``a dataset that this technique has not yet been evaluated on'' requires also looking at the dataset-evaluation history of other techniques that address the same task. This relation does not appear within the one-hop range and appears only within the two-hop range (the four specific kinds of graph gaps are covered in Stage 5 of Section~\ref{sec:pipeline}).

\subsection{Pipeline: Nine Stages}
\label{sec:pipeline}

ShadowMiner processes the collected documents in the following nine stages, from document-importance re-ranking (Stage 1) to the citation-faithfulness check (Stage 9).

\begin{figure}[tb]
\centering
\includegraphics[width=0.95\linewidth]{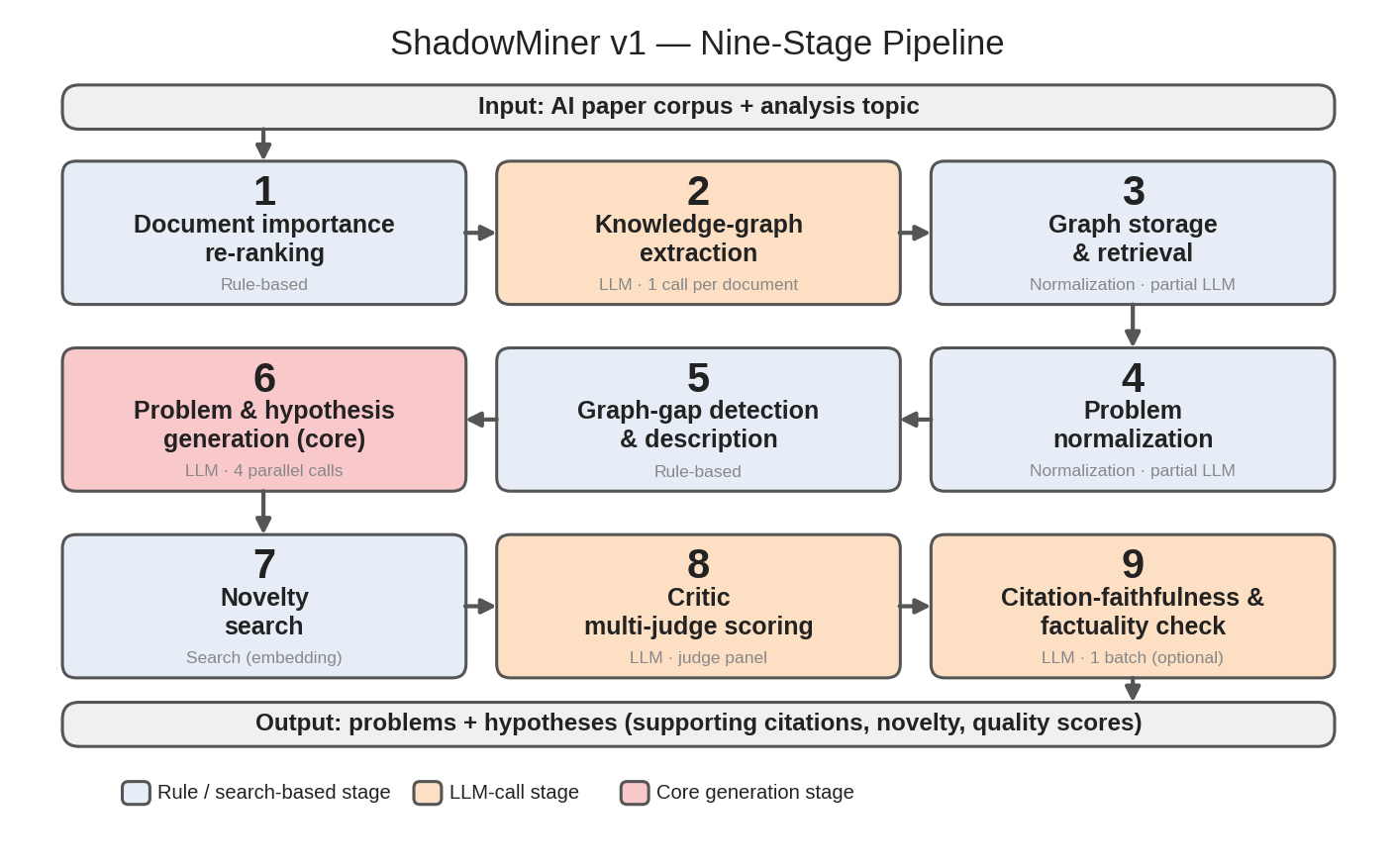}
\caption{Nine-stage pipeline, from document-importance re-ranking to the citation-faithfulness check. Stages 2, 6, 8, and 9 call an LLM throughout; Stages 3 and 4 call an LLM only in part, during normalization; the rest are rule- or embedding-based.}
\label{fig:pipeline}
\end{figure}

Whereas Figure~\ref{fig:pipeline} shows the overall flow and whether each stage calls an LLM at a glance, Table~\ref{tab:stages} summarizes what each stage does, one line per stage and in more concrete terms.

\begin{table}[tb]
\caption{Name, LLM use, and one-line description of each of the nine stages.}
\label{tab:stages}
\centering
\footnotesize
\begin{tabular}{cL{3.2cm}L{2.0cm}L{6.5cm}}
\toprule
\# & Stage & LLM calls & Description \\
\midrule
1 & Document-importance re-ranking & None & Adds a small weight from external scholarly metadata (citation count, field-disruption index) to search relevance to set the order of the documents used as evidence \\
2 & Knowledge-graph extraction & One per document (parallel) & Calls an LLM once per document to extract concepts and relations that fit a fixed schema \\
3 & Graph storage and retrieval & Partly, during normalization & Accumulates the extracted facts in a store and builds two kinds of graph, narrow and wide \\
4 & Problem normalization & Partly & Merges problems that differ only in wording so that they are not counted as separate limitations \\
5 & Graph-gap detection and description & None & Uses rules on the wide graph to find graph gaps and turns each into one described line that is passed to Stage 6 \\
6 & \textbf{Problem and hypothesis generation} & Parallel calls (default 4) & Sends the narrow graph and the hints to the LLM in parallel with the same prompt and merges the results into problems and hypotheses \\
7 & Novelty search & None & Compares each generated hypothesis with existing papers by semantic (embedding) distance to gather candidates \\
8 & Critic multi-judge scoring & Judge panel & Scores each hypothesis on inclusion decision, novelty, and four-axis quality \\
9 & Citation-faithfulness and factuality check (optional) & One batched call & Checks whether the generated statements match the cited evidence \\
\bottomrule
\end{tabular}
\end{table}

Each stage in Figure~\ref{fig:pipeline} and Table~\ref{tab:stages} is described in more detail below.

\paragraph{Stage by stage.}

\paragraph{1. Document-importance re-ranking} --- Documents are ordered by search relevance (normalized to 0--1) plus 0.15 times an importance score (0--1). The importance score is computed from citation count and the disruption index [47], both taken from the external dataset SciSciNet [45]. The disruption index looks at whether papers that cite a given paper also cite the earlier references that the given paper itself cited. The more often later citations point to the paper without those earlier references, the more the paper is regarded as replacing earlier work, and the index is positive; the more often they cite it together with its earlier references, the more it is regarded as building on earlier work, and the index is negative. In how the score is applied, citation count (on a log scale) accounts for most of it, and disruption adds up to 0.3 only when it is positive (there is no penalty when it is negative). Only documents with approximately tied relevance are reordered; when the relevance difference is large, the order is not reversed.

\paragraph{2. Knowledge-graph extraction} --- The first 12,000 characters of each document body are given to the LLM, which extracts entities and triples using the six entity types and eight relation types of Section~\ref{sec:system}. Extracted entities and relations of any other type are discarded. Documents are processed in parallel. If extraction fails for one document, only that document is skipped and the rest continue (a warning is logged if the failure rate exceeds 50\%).

\paragraph{3. Graph storage and retrieval} --- The concept names in the extracted triples are first normalized, meaning that different names for the same concept are merged into one; the triples are then accumulated in a persistent store. Normalization is a three-step procedure: first, pairs of names with somewhat similar strings are selected as candidates; second, candidates whose strings are nearly identical or whose embeddings are very close are merged; third, an LLM judges, for each remaining candidate, whether the two names refer to the same concept. Accumulation must be persistent because graph-gap detection (Stage 5) requires several papers that share a dataset, a task, or a limitation, and the input papers alone do not include enough such papers. GoR [15], EvoNarrator [12], IDEAgent [21], and others accumulate the genealogy of the generated ideas (hypotheses) themselves across sessions; ShadowMiner, in contrast, accumulates only triples and does not remember the hypotheses across sessions (item 7 of Table A1).

\paragraph{4. Problem normalization} --- Different papers can write the same limitation in different words (e.g., ``lack of X'' and ``insufficient X''), and these would otherwise be counted as separate limitations. This stage merges Problem entities that are written differently but refer to the same problem, among those connected by the relations limited-by (``pointed out as a limitation'') and addresses-problem (``claims to solve this problem''). The merging uses the same three-step procedure as Stage 3.

\begin{figure}[tb]
\centering
\includegraphics[width=0.95\linewidth]{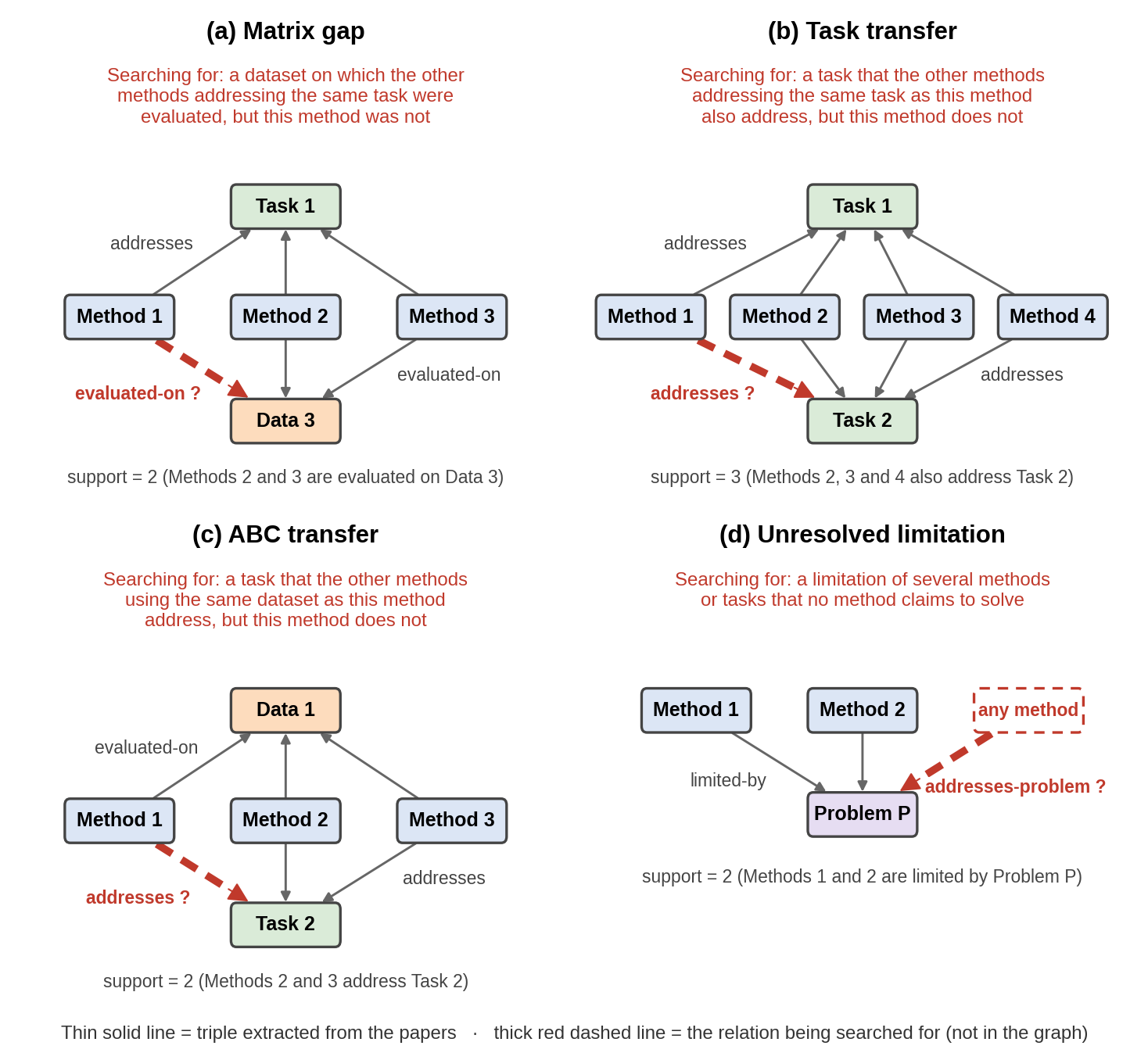}
\caption{The four kinds of graph gaps. In each panel, thin solid lines are triples extracted from the papers, and the thick red dashed line is the relation being searched for, which is not in the graph. (a) Matrix gap: Methods 2 and 3, which address the same Task 1, are evaluated on Data 3, but Method 1 is not (support 2). (b) Task transfer: Methods 2, 3 and 4 address Task 2 in addition to Task 1, but Method 1 does not (support 3). (c) ABC transfer: Methods 2 and 3, which use the same Data 1, address Task 2, but Method 1 does not (support 2). (d) Unresolved limitation: Methods 1 and 2 are limited by Problem P, and no method is recorded as claiming to solve it (support 2). The methods, tasks, datasets, and problem in the panels are placeholders.}
\label{fig:graphgaps}
\end{figure}

Figure~\ref{fig:graphgaps} shows, for each of the four kinds, which relation is searched for and how support is counted.

\paragraph{5. Graph-gap detection and description} --- This stage finds graph gaps in the wide graph by rule-based computation. A graph gap is either a relation that similar items have in the triples extracted from the papers we read but one item lacks, or a limitation that is recorded for several methods or tasks but that no method is recorded as claiming to solve. There are four kinds of graph gaps: matrix gap, task transfer, ABC transfer, and unresolved limitations (Figure~\ref{fig:graphgaps}). The first three work in the same way (Table~\ref{tab:patterns}): methods are grouped by the task they address or the dataset they use, and what the other methods in the group have but one method lacks is reported as a graph gap. The fourth kind, unresolved limitations, works differently: it finds limitations that are recorded for several methods or tasks and that no method is recorded as claiming to solve. Each graph gap found is written as a one-line text string, called a described graph gap (DGG). For example, \texttt{[matrix gap] 'Method 1' addresses 'Task 1' but is not evaluated on 'Data 3' (2 methods of that task use 'Data 3') $\rightarrow$ comparison possible when evaluated} (an illustrative example, translated from the Korean text that the system produces). DGGs are sorted by support (the number of supporting methods) and given to the LLM in Stage 6.

\begin{table}[tb]
\caption{Three matching patterns of graph-gap detection --- the criterion for grouping methods, and what only this method lacks.}
\label{tab:patterns}
\centering
\begin{tabular}{L{4.2cm}L{3.8cm}L{4.2cm}}
\toprule
Pattern & Grouping criterion & What only this method lacks \\
\midrule
(1) Matrix gap (application of Zwicky's morphological analysis [40]) & Methods that address the same task & Datasets used by the other methods in the same group \\
(2) Task transfer & Methods that address the same task & Additional tasks addressed by the other methods in the same group \\
(3) ABC transfer (application of Swanson's ABC linking [36]) & Methods that use the same dataset & Tasks addressed by the other methods in the same group \\
\bottomrule
\end{tabular}
\end{table}

\paragraph{6. Problem and hypothesis generation (core stage)} --- The narrow graph, the described graph gaps (DGGs), and the limitation summary are combined into one prompt and sent to the LLM (by default the same prompt is sent as four parallel calls and the results are merged). The limitation summary is a list of the top 20 problems that the paper authors themselves pointed out as limitations, in descending order of how many times they were pointed out, regardless of whether they have been resolved. The DGGs already include the unresolved limitations found in Stage 5. The limitation summary is added separately so that the prompt also contains limitations that the DGGs do not include, such as those recorded as solved. Based on this input, the LLM derives research problems and, for each problem, produces one or two hypotheses, each consisting of a prediction (``doing A will improve B''), an expected result, and a verification method, together with supporting citations.

\paragraph{7. Novelty search} --- Using the prediction of each generated hypothesis as the query, we run hybrid search over the paper corpus (keyword search and vector search combined with RRF, Reciprocal Rank Fusion) and attach the title, abstract, and year of the top three results as ``similar existing work.'' Graph-gap detection (Stage 5) considers only the documents collected for a given topic, so a gap identified there may already be addressed by work present in the full corpus.

\paragraph{8. Critic multi-judge scoring} --- Several judges (composed by mixing different LLM providers) evaluate each hypothesis on three things. The \textbf{inclusion decision}: a hypothesis is dropped from the final result if it lacks evidence, is trivial, cannot be verified, or is unrelated to the topic. \textbf{Novelty}: exists, new, or unclear. \textbf{Four-axis quality scores}: soundness, verifiability, groundedness (whether the cited evidence, the triples and the similar existing work, actually supports the hypothesis), and topical relevance, each on a scale of 1 to 5. When there are several judges, their results are combined for each hypothesis: the inclusion decision and novelty take the value chosen by most judges (majority vote), and each of the four-axis scores takes the median. When a judge gives no result for a hypothesis, the results of the other judges are combined by the same rules. If the keep/drop votes are evenly split, the hypothesis is included. If the novelty votes are evenly split, the result defaults to \texttt{glm-5.1}.

\paragraph{9. Citation-faithfulness and factuality check} --- An LLM checks two things. \textbf{Citation faithfulness} looks at whether a problem or a hypothesis's prediction, as generated in Stage 6, is actually supported by the documents it cites. \textbf{Factuality} is finer-grained: it splits that text into the smallest indivisible factual units (``atomic facts'') and checks whether each is supported. For example, ``A achieved 90\% accuracy on B'' is split into ``A addresses B'' and ``the accuracy is 90\%.''

\subsection{Design Rationale}
\label{sec:rationale}

\paragraph{Why a knowledge graph.} Instead of feeding the original text of papers into the LLM as it is, we chose to structure it as a graph first. To find ``combinations this technique has not yet tried'' by rule (Stage 5 of Section~\ref{sec:pipeline}), concepts from different papers must be linkable, and this is hard to do reliably by rereading the original text each time. The approach of finding missing combinations in a knowledge graph in order to generate hypotheses already exists as a line of research that uses link prediction [46]; ShadowMiner fills this role with threshold-based rule computation instead of a learned prediction model. There is also a design that does not use a knowledge graph and hands paper (title, abstract) pairs directly to the LLM, letting the LLM do all of the retrieval, composition, and ranking (MOOSE-Chem [26]). We do not compare performance directly with this design: even if we did, the results could vary with the version of the LLM used, which would make them hard to interpret as a difference in the designs themselves.

\paragraph{Why an LLM as an evaluation tool.} Novelty judgment, critic scoring, and the citation-faithfulness check all use an LLM. This is not because an LLM is essential for these judgments, but because we have not yet had human evaluators review them, so an LLM was used as a substitute. How far this judging tool can be trusted must therefore be checked separately, which Section~\ref{sec:reliability} covers.

\section{Techniques and Their Sources}
\label{sec:techniques}

Table A1 in Appendix~\ref{app:techniques} lists 13 techniques for ShadowMiner, whether taken from other work or devised by ourselves. For each technique, it gives its source (what was taken from where), its implementation, its measured results, and whether it was finally adopted. Every technique from other work carries its source regardless of adoption, and our own improvements are marked as such.

\section{Evaluation}
\label{sec:eval}

\subsection{Setup}
\label{sec:setup}

\paragraph{Evaluation scope.} ShadowMiner, as introduced in this report, takes AI papers as input and discovers research problems and hypotheses. All the experiments below were run under exactly this definition, on a corpus of research papers only.

\paragraph{Contamination boundary.} In the actual-appearance experiment (Section~\ref{sec:mainresults} below), we took the release date of the generation model (Zhipu AI's GLM-family \texttt{glm-4.7}, 2025-12-22) as the basis and set 2026-01-01 as the reference time for search and judgment (freeze date = contamination boundary). With this setting, every candidate answer is a paper published after the freeze date, so it could not have been in the generation model's training data. This date boundary is the only contamination control used in this study; Section~\ref{sec:discussion} discusses its limitations.

\paragraph{Judging model.} The actual-appearance judgment is made by OpenAI's reasoning model \texttt{o3}. For each pair of a hypothesis and a candidate paper, \texttt{o3} is given the hypothesis together with the paper's title and abstract and is asked whether the paper takes the same core approach to the same problem as the hypothesis; the specific methods do not have to be identical. In initial trials, \texttt{o3} did not always give the same answer to the same question, so we ask the same question three times for each pair and accept the answer given by at least two of the three (three-vote majority).

\paragraph{Why we did not use a public external benchmark.} We examined AI Idea Bench 2025 [48], the public benchmark that seemed most suitable for our evaluation, but did not adopt it. This benchmark gives a system only the five papers cited by a target paper, asks it to guess the idea of that target paper, and then scores the result by comparing it with the target paper's actual idea. Because the score is how closely a system's idea matches the target paper's idea, an idea that is not in the target paper receives a low score even if other papers contain it. There is also a contamination problem. The target papers in this benchmark were published after GPT-4o's knowledge cutoff (2023-10-03), but our generation model (\texttt{glm-4.7}) was released about two years and two months later, on 2025-12-22, so those target papers may already be included in the LLM's training data. There is also a fairness problem. The other systems already scored on this benchmark all obtained their scores by following the rule of five cited papers as input, whereas ShadowMiner is built to gather far more documents and construct a knowledge graph from them. If we gave it only five papers as the rule requires, the graph would be sparse and ShadowMiner's ability would be understated; conversely, if we expanded the documents for ShadowMiner alone, it would receive an advantage that the other systems did not --- either way, its score could not be compared side by side with those of the other systems. There is also a base-model mismatch: the systems we would compare against all used the same model (GPT-4o-2024-11-20), while ShadowMiner's generation model is different, so a difference in scores could come from the model rather than the method.

\paragraph{Topic composition.} The targets are nine AI/CS topics (\texttt{efficient\_inference}, \texttt{rag\_hallucination}, \texttt{alignment}, \texttt{agents}, \texttt{multimodal}, \texttt{code\_generation}, \texttt{reasoning}, \texttt{small\_lm}, \texttt{eval\_benchmark}). For all nine topics the pipeline ran to the end and produced results (9/9 completed), and no run stopped midway or ended in an error (0 failures). The median time per topic was 91 minutes (range 65--118 minutes).

\subsection{Main Results --- Actual Appearance Rate}
\label{sec:mainresults}

\paragraph{Method.} We roll ShadowMiner's clock back to 2026-01-01 (the ``freeze'') and have it generate hypotheses from only the literature up to that date. To see whether each hypothesis later appeared, we use its prediction and verification method as the search query and retrieve up to six candidate papers published after the freeze date. Starting with the top-ranked candidate, we ask \texttt{o3} whether each of the five highest-ranked papers matches the hypothesis (its prediction); each paper is judged by a three-vote majority (Section~\ref{sec:reliability}). If any of the five is judged a ``match,'' we stop there and count the hypothesis as having appeared; if all five are judged non-matches, it is recorded as not having appeared. In other words, we measure whether a hypothesis that ShadowMiner produced actually appeared in research published after the freeze date.

\begin{table}[tb]
\caption{Actual appearance results by topic (all nine topics). ``Problem-level appearance'' is the proportion of cases in which the problem statement itself --- the stage above the hypothesis, not the hypothesis --- later matched actual research.}
\label{tab:table4}
\centering
\small
\begin{tabular}{lccC{2.1cm}C{2.1cm}}
\toprule
Topic & Hypotheses & Appeared & Actual appearance rate & Problem-level appearance \\
\midrule
efficient\_inference & 12 & 2 & 0.167 & 5/6 \\
rag\_hallucination & 12 & 3 & 0.250 & 3/6 \\
alignment & 10 & 3 & 0.300 & 2/6 \\
agents & 11 & 6 & 0.545 & 6/6 \\
multimodal & 12 & 8 & 0.667 & 4/6 \\
code\_generation & 12 & 7 & 0.583 & 5/6 \\
reasoning & 12 & 5 & 0.417 & 6/6 \\
small\_lm & 10 & 3 & 0.300 & 6/6 \\
eval\_benchmark & 12 & 3 & 0.250 & 4/5 \\
\midrule
Total & 103 & 40 & 0.388 & 0.774 (41/53) \\
\bottomrule
\end{tabular}
\end{table}

The most representative and important result, the actual appearance rate, is 0.388 (40/103) (Table~\ref{tab:table4}). The Wilson 95\% confidence interval (CI), treating the 103 hypotheses as independent samples, is [0.300, 0.485], but in reality hypotheses within the same topic are not independent. The evidence is that appearance rates differ greatly across topics (0.167--0.667): if the hypotheses were truly independent, each topic's rate would lie much closer to the overall rate of 0.388, so the large differences show that hypotheses from the same topic share conditions specific to that topic and are correlated with one another. Recomputing with a bootstrap that resamples whole topics (with replacement, the nine topics as units) gives [0.282, 0.500], which is 18\% wider --- the independence assumption was optimistic. The actual appearance rate of 0.388 is therefore a rough guide, not an exact figure: the true rate could be lower or higher, plausibly anywhere from about 28\% to 50\%.

\begin{figure}[tb]
\centering
\includegraphics[width=0.95\linewidth]{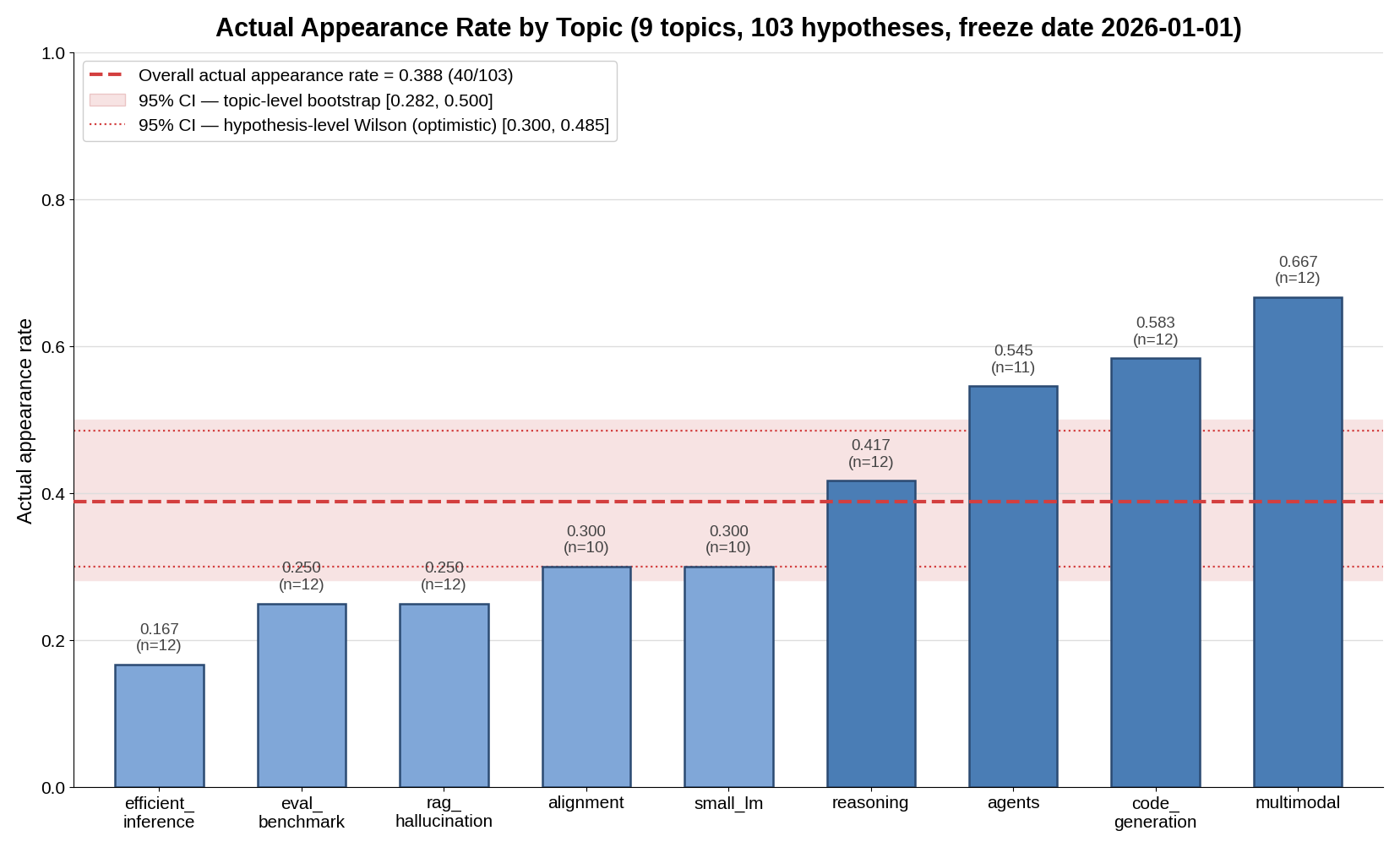}
\caption{Actual appearance rate for the nine topics (efficient\_inference 0.167 to multimodal 0.667). The thick red dashed line is the overall actual appearance rate (0.388), the light shading is the topic-level bootstrap 95\% CI ([0.282, 0.500], the interval we adopt), and the dotted lines are the hypothesis-level Wilson CI (independence assumed, optimistic). The variation across topics is the direct cause of the difference in width between the two CIs.}
\label{fig:mainresults}
\end{figure}

Figure~\ref{fig:mainresults} shows visually the variation across topics in Table~\ref{tab:table4} and the resulting difference in width between the two confidence intervals. In the end, which topics are chosen strongly affects the actual appearance rate, and this single value (0.388) does not reveal that dependence on topic.

\paragraph{Source composition.} The actual-appearance judgment checks two things: whether a candidate paper matches the hypothesis, and whether its publication date is after the freeze date (2026-01-01). The candidate papers come from four sources (arxiv, openalex, acl, openreview), and the publication date has a different meaning in each. In arxiv it is the v1 submission date, which is the date the content was first made public. In openalex it is the journal publication date and in acl the proceedings publication date, and either can be later than the date the content was first made public. This means some papers counted as published after the freeze date had, in fact, already been made public before it. Such a paper may have been in the training data of the generation model. Of the 40 hypotheses that appeared, that is, matched a paper published after the freeze date, 27 matched an arxiv paper, 12 an openalex paper, 1 an acl paper, and none an openreview paper. Counting as appeared only the 27 hypotheses that matched an arxiv paper --- whose recorded date is always its first public date --- gives a rate of 0.262 (27/103). This value is a lower bound: we did not check whether the 13 hypotheses matched to a paper from another source would also match a paper on arxiv. If some of them would, the true value would be higher than 0.262. We report it together with the actual appearance rate of 0.388 (40/103).

\paragraph{Actual appearance rate by novelty.} In Stage 8, the critic panel assigns each hypothesis a novelty label by comparing it with the similar papers found by the novelty search of Stage 7: ``exists'' if a similar paper already covers the hypothesis, ``new'' if none does, and ``unclear'' otherwise. We compared how often hypotheses with each label appeared. Of the 21 hypotheses labeled ``exists,'' 12 appeared (0.571); of the 71 labeled ``new,'' 24 appeared (0.338); the difference is +0.233 ($p=0.054$). The hypotheses that the critic panel judged to be less novel therefore appeared more often in later papers. These numbers are computed on 92 of the 103 hypotheses: 3 are excluded because they are labeled ``unclear,'' and the other 8 are excluded because of a leak that occurred while Stage 7 was enforcing the freeze date. Because the novelty label depends on the papers that Stage 7 retrieves, it inherits a limitation in how Stage 7 enforces the freeze date. Stage 7 applies the freeze date to the papers it retrieves, but when a paper's source records only its publication year, not an exact date, we default it to January 1 of that year --- the same date as the freeze date --- so it can still be included even if it was actually published later in the year. This affected 8 of the 103 hypotheses. We removed these 8 hypotheses and computed on the remaining 92 that are labeled ``exists'' or ``new'' (Appendix~\ref{app:robustness}). To check whether excluding these 8 hypotheses changed the result, we also computed it including them, on all 100 hypotheses labeled ``exists'' or ``new'' (excluding the 3 labeled ``unclear''): 0.538 (14/26) against 0.338 (25/74), a difference of +0.201 ($p=0.071$) --- the same direction. We do not know why the hypotheses labeled ``exists'' appeared more often. One possible explanation, which we have not tested, is this: a hypothesis labeled ``exists'' indicates that its research area was already active before the freeze date; an area that is already active is likely to keep producing papers, so a hypothesis in that area is more likely to be matched by one of them after the freeze date too. When an independent judge model (\texttt{o3}), which was not on the panel, assigned the labels again, the difference had the same direction but was not statistically significant ($p=0.130$; Section~\ref{sec:reliability}).

\paragraph{Self-assessment has no predictive power.} In Stage 8, the critic panel gives each hypothesis an overall quality score (the self-assessment score). The overall score is the mean of five scores: soundness, verifiability, groundedness, topical relevance, and novelty (new = 5, unclear = 3, exists = 1). For each topic, we compared the mean self-assessment score of the hypotheses that appeared (matched a paper published after the freeze date) with that of the hypotheses that did not appear. In some topics the hypotheses that appeared have higher self-assessment scores, while in others those that did not appear have higher scores (the per-topic difference between the two means ranges from $-0.175$ to $+0.377$), so the self-assessment score alone cannot tell whether a hypothesis will appear.

The differences between the two means are positive in some topics and negative in others. We checked whether these differences could be random variation rather than a real pattern. In each topic the hypotheses are split into two groups, those that appeared and those that did not, and the difference is between the means of these two groups. The difference is unreliable if either group is small, so what matters is the number of hypotheses in the smaller group (the size of the smaller group), not the number in the whole topic: a topic with 12 hypotheses, of which 2 appeared and 10 did not, has a smaller group of size 2. If the differences were random variation from small groups, they would tend to be larger in topics where the smaller group is smaller. Across the nine topics, the correlation between the size of the smaller group and the magnitude of the difference (ignoring its sign) is $r=0.049$ ($p=0.901$) --- essentially no relationship, giving no support for this explanation. Nine topics are not enough to tell whether the differences are random variation or a real pattern.

\paragraph{Reference upper bound.} As a reference, we ran the same experiment with the same generation model and settings, changing only the freeze date to 2024-06-01 instead of 2026-01-01. In this run, a hypothesis counts as appeared if it matches a paper published after 2024-06-01, a span of about 27 months, which is much longer than in the main experiment. With a different freeze date, the input literature is different, so we generated a new set of hypotheses for this run; it contains 87 hypotheses rather than 103. The actual appearance rate in this run is 0.506 (44/87; Wilson 95\% CI [0.403, 0.608]). The generation model was released on 2025-12-22, so it may have been trained on papers published between 2024-06-01 and that date. In this run, a hypothesis may match a paper because the generation model had learned that paper (memorization) rather than through a genuine prediction, and we cannot tell the two apart. We report 0.506 only as a reference value obtained under a condition where memorization is possible, and we do not use the difference between 0.388 and 0.506 as evidence about ShadowMiner's performance.

\begin{table}[tb]
\caption{Comparison of three values of the actual appearance rate.}
\label{tab:table5}
\centering
\begin{tabular}{L{4.5cm}cL{5.5cm}}
\toprule
Appearance rate & Value & Condition \\
\midrule
Conservative lower bound (arxiv-confirmed matches only) & 0.262 (27/103) & Only arxiv, whose recorded date is its first public date \\
Actual appearance rate & 0.388 (40/103) & All four sources; contamination boundary = generation model release date \\
Reference upper bound (freeze date moved to 2024-06-01) & 0.506 (44/87) & Memorization possible; separate run \\
\bottomrule
\end{tabular}
\end{table}

\subsection{Reliability of the Judging Tool}
\label{sec:reliability}

We have not yet verified the judgments by human scoring (Section~\ref{sec:discussion}). Instead, we ran checks on each of the two LLM judgments used in Section~\ref{sec:mainresults}. The actual-appearance judgment, made by \texttt{o3}, produces the actual appearance rate (0.388). We checked it with judgment agreement, a negative control, and a cross-model check. The novelty judgment, made by the critic panel, produces the novelty labels (exists/new/unclear). We checked it with a similarity check and a cross-model relabeling.

\paragraph{Judgment agreement.} For each pair of a hypothesis and a candidate paper, the actual-appearance judgment asks \texttt{o3} the same question three times and accepts the majority answer (Section~\ref{sec:setup}). We call a match unanimous if all three answers are ``yes'' and a split verdict if two are ``yes'' and one is ``no.'' With a split verdict, \texttt{o3} did not give the same answer every time, so the result might differ if the question were asked again. Of the 40 hypotheses that appeared, 33 (82.5\%) were matched unanimously and 7 (17.5\%) by a split verdict. For these 7 hypotheses, therefore, one of the three answers of \texttt{o3} was ``no.''

\paragraph{Negative control.} A negative control measures how often the judge answers ``match'' when the correct answer is ``no match'' (the false-positive rate). We checked each of the 103 hypotheses against papers of an unrelated topic. For this, each of the nine topics was assigned one other topic of unrelated content, and the hypotheses of the topic were checked against papers retrieved for the assigned topic; for example, the hypotheses of \texttt{agents} were checked against papers retrieved for \texttt{eval\_benchmark}. The papers were retrieved by searching for papers published after the freeze date, using the description of the assigned topic as the query, and \texttt{o3} (three-vote majority, as in the actual-appearance judgment) judged whether any of the top five retrieved papers matches the hypothesis. Because the assigned topic is unrelated to the topic of the hypothesis, the correct answer is ``no match'' in every case. We tried to choose pairs with no known overlapping subfield, but did not fully avoid it: two pairs may still overlap, \texttt{rag\_hallucination} with \texttt{agents} (agentic retrieval-augmented generation) and \texttt{small\_lm} with \texttt{reasoning} (reasoning in small models). If a retrieved paper is in fact related to a hypothesis, an answer of ``match'' would be correct, but because this experiment treats ``no match'' as the correct answer in every case, we would still count it as a false positive. The false-positive rate measured here may therefore be higher than the true rate. Across the nine topics, one hypothesis was judged to match a retrieved paper (1/103; Wilson 95\% CI [0.002, 0.053]). The match came from the pair of \texttt{eval\_benchmark} and \texttt{rag\_hallucination}: a hypothesis about filtering contaminated retrieval results in search-based agents was judged to match a paper about filtering inconsistent retrieved documents in retrieval-augmented generation. This pair was not one of the two we flagged above as possibly overlapping, but the hypothesis is itself about retrieval contamination filtering, so the two topics may overlap here as well.

\paragraph{Cross-model check.} The three votes of the actual-appearance judgment all come from the same model, \texttt{o3}, so a bias of that model would affect all three votes. We therefore asked a different model, \texttt{gpt-5}, to judge again the same 40 pairs of a hypothesis and its matched paper --- the 33 matches with \texttt{o3}'s unanimous votes and the 7 with \texttt{o3}'s split votes --- without showing it the answers of \texttt{o3}. \texttt{gpt-5} also answered by a three-vote majority. Of the 33 matches with \texttt{o3}'s unanimous votes, \texttt{gpt-5} judged 14 to be a match unanimously, 4 to be a match by a split vote, and the remaining 15 to be no match. Of the 7 matches with \texttt{o3}'s split votes, \texttt{gpt-5} judged all 7 to be no match. Without human judgments, we cannot tell which model is correct where they disagree, so the actual appearance rate should be read as a value that depends on the judging model, especially for the split verdicts.

Taken together, the three checks (negative control, judgment agreement, and cross-model check) show how far the actual-appearance judgment can be trusted. \texttt{o3} rarely judged an unrelated paper to be a match in the negative control (1 of 103). For the actual-appearance judgment, however, \texttt{o3}'s judgments were less consistent: 7 of the 40 matches were split verdicts, and 22 of the 40 were not confirmed by \texttt{gpt-5}. The actual appearance rate (0.388) therefore includes matches on which the judging models disagree.

To check how far the critic panel's novelty labels (exists/new, assigned in Stage 8) can be trusted, we ran a similarity check and a cross-model relabeling. When the critic panel assigns a novelty label to a hypothesis, it is given the hypothesis and the three papers that the novelty search of Stage 7 found as ``similar existing work,'' and it decides whether these papers already cover the hypothesis. In the similarity check, we measured how similar each hypothesis is to these three papers, as the highest embedding cosine similarity between the hypothesis and the abstracts of the papers. If the critic panel's labels are valid, a hypothesis labeled ``exists'' should be more similar to these papers than a hypothesis labeled ``new.'' The mean over the 103 hypotheses is 0.657 for the 26 labeled ``exists,'' 0.613 for the 74 labeled ``new,'' and 0.617 for the 3 labeled ``unclear''; the difference between ``exists'' and ``new'' is small, so this check does not clearly confirm the expectation. In the cross-model relabeling, \texttt{o3}, which was not a member of the critic panel, labeled all 103 hypotheses again with the same similar papers. Its labels agree with the panel's for 68 of the 103 (0.660), but this agreement is uneven: of the 74 hypotheses the panel labeled ``new,'' \texttt{o3} agreed on 66; of the 26 the panel labeled ``exists,'' \texttt{o3} agreed on only 2. The two models agreed far more often on ``new'' than on ``exists.'' Because the Section~\ref{sec:mainresults} finding --- that hypotheses labeled ``exists'' appeared more often than those labeled ``new'' --- depends on this less stable ``exists'' label, that finding may also change depending on which model assigned the labels. We then recomputed the Section~\ref{sec:mainresults} appearance-rate difference between ``exists'' and ``new'' hypotheses, using \texttt{o3}'s labels instead of the panel's; the result is in Table~\ref{tab:table6}.

\begin{table}[tb]
\caption{Exists/new appearance-rate difference by the judge that assigned the novelty labels.}
\label{tab:table6}
\centering
\small
\begin{tabular}{lC{1.9cm}C{1.9cm}cC{2.4cm}}
\toprule
Label basis & Exists appearance rate & New appearance rate & Difference & Significance \\
\midrule
Original three-member panel & 0.538 (14/26) & 0.338 (25/74) & +0.201 & $z=1.80$, $p=0.071$ \\
Independent relabel (o3) & 0.750 (3/4) & 0.372 (29/78) & +0.378 & $z=1.51$, $p=0.130$ \\
\bottomrule
\end{tabular}
\end{table}

The ``Original three-member panel'' row is the same data already given in Section~\ref{sec:mainresults}: the panel's own appearance rates on the 100 hypotheses it labeled ``exists'' or ``new'' (it labeled the other 3 ``unclear''). The ``Independent relabel'' row shows the result of \texttt{o3} making the same judgment: \texttt{o3} labeled 21 of the 103 hypotheses ``unclear,'' leaving 82. As Table~\ref{tab:table6} shows, the appearance rate of hypotheses \texttt{o3} labeled ``exists'' (0.750) is still higher than that of those it labeled ``new'' (0.372) --- the direction is maintained. \texttt{o3} labeled far fewer hypotheses ``exists'' (4, versus the panel's 26), so this appearance rate is estimated from just 4 cases. A rate estimated from so few cases is imprecise, with a wide confidence interval --- which is why, even though the difference between the two groups grew larger (+0.378 vs. +0.201), the $p$-value rose from 0.071 to 0.130, and the difference no longer meets the commonly used significance criterion ($p<0.05$).

\section{Negative Results}
\label{sec:negative}

This section reports negative results: techniques we tried, and a check on the technique's predictive benefit and limitation. Most of the comparisons below reuse three of the nine topics from the main experiment of Section~\ref{sec:mainresults}, using the same documents and knowledge graph that were given to each. The three topics are multimodal models, RAG hallucination, and inference efficiency (quantization), which differ in nature. For each comparison, we kept each topic's documents and knowledge graph fixed, changed only the variable under test, and reran generation and verification; we also recomputed the actual-appearance rate of the resulting hypotheses, using the same method as Section~\ref{sec:mainresults}. The LLM can give different outputs each time it is run on the same input, and each comparison covers only a small number of hypotheses --- about a dozen per topic. We therefore treat a small difference with caution, because it may come from the non-determinism of the LLM rather than from the changed variable.

\paragraph{Generation mode --- sequential dropped.} We compared two ways of calling the LLM N times to generate hypotheses. In the parallel mode, the N calls are made independently at the same time with the same prompt, and the results are merged and deduplicated. In the sequential mode, the calls are made one after another; from the second call on, the prompt includes the problems proposed so far, together with a one-line summary of each problem's hypotheses, and instructs the LLM to propose a clearly different approach, or to frame the problem differently. In the comparison on the three topics (Table A1, item 3), the sequential mode had a higher waste rate (the extent to which generated hypotheses were discarded as duplicates) than the parallel mode in every topic (0.63--0.71 against 0.46--0.63) and took 1.0 to 1.8 times as long, while the mean groundedness score was about the same (sequential 3.05 against parallel 3.03). The actual appearance rate was lower with the sequential mode (5/24 = 0.208) than with the parallel mode (12/31 = 0.387), consistent with the waste rate and speed results above. Though the sample is small and the difference is not statistically significant ($p=0.155$), we kept the parallel mode as the default, based on the waste rate, the speed difference, and this appearance rate.

\paragraph{Attaching sources to graph gaps.} By default, the described graph gap (DGG; Section~\ref{sec:pipeline}) produced by Stage 5 is a one-line string with three parts: the kind of the graph gap, a sentence stating what is missing, and a direction phrase that is the same for every graph gap of that kind (an illustrative example, translated from the Korean text that the system produces).

\begin{quote}
\texttt{[matrix gap] 'Method 1' addresses 'Task 1' but is not evaluated on 'Data 3' (2 methods of that task use 'Data 3') $\rightarrow$ comparison possible when evaluated}
\end{quote}

This line does not say which papers the graph gap was derived from, except for the unresolved-limitation kind. The LLM must therefore find the supporting evidence itself and relate it to the graph gap. To test whether giving this information helps, we attached the numbers of the source papers to the DGG. We call the result the described graph gap with source papers (DGG-SP). It contains the numbers of the papers from which the triples that establish the graph gap were extracted (up to six) and, when there is one, the unresolved limitation of the method in the graph gap (example, translated).

\begin{quote}
\texttt{Kind: matrix gap} \\
\texttt{Gap: 'Method 1' addresses 'Task 1' but is not evaluated on 'Data 3' (2 methods of that task use 'Data 3')} \\
\texttt{Source-paper numbers: [3][7][12]} \\
\texttt{Linked unresolved limitation: degraded long-context performance} \\
\texttt{Research direction: comparison possible when evaluated}
\end{quote}

A prior study (Agentic-Ideation [1]) found that an explicit gap-analysis step (\texttt{Analyse\_Gap}) raised the novelty score from 5.86 to 6.74 (Table A1, item 10), suggesting that making this step explicit helps. Taking this idea, we tested whether giving the LLM the evidence for each graph gap --- the DGG-SP --- affects the hypotheses it produces. The mean groundedness score across the three topics, assigned by the critic (on a scale of 1 to 5), was about the same for the DGG and the DGG-SP (mean 3.24 against mean 3.22). By topic, it rose with the DGG-SP in multimodal (3.17 to 3.50), stayed about the same in RAG hallucination (3.25 to 3.25), and fell in inference efficiency (3.30 to 2.92). The actual appearance rate was higher with the DGG-SP than with the DGG (17/36 = 0.472 against 14/34 = 0.412), but the difference was not statistically significant ($p=0.611$). The longer prompt of the DGG-SP also increased the cost in every topic. We found no measurable benefit that offsets this added cost, and therefore kept the DGG as the default.

\paragraph{Critic-feedback regeneration loop.} This loop regenerates a hypothesis that received a weak evaluation in the critic scoring (Stage 8) or the citation-faithfulness check (Stage 9). The LLM is given the original hypothesis and the reason for the weak evaluation. The new hypothesis replaces the original only if that evaluation improves. Stage 9 (Section~\ref{sec:pipeline}) gives each hypothesis a citation-faithfulness grade of ``full,'' ``partial,'' or ``none,'' according to how far the documents it cites support it. Hypotheses are selected by two criteria applied in order. First, hypotheses with the grade ``none'' or ``partial'' are selected, those with ``none'' first. Only if there is no such hypothesis, hypotheses with a groundedness or verifiability score below 3 out of 5 (Stage 8) are selected. At most four hypotheses are regenerated per topic. Only the selected hypothesis is regenerated, not the whole set of hypotheses from that topic. The LLM receives the same input as in Stage 6 (the narrow graph, the DGGs, and the limitation summary) and, in addition, the original hypothesis, the reason it was selected, and the other hypotheses from that topic, so that the new hypothesis does not repeat them. The new hypothesis replaces the original only if it meets three conditions: it is not a near-duplicate of an existing hypothesis (using the same similarity threshold as the deduplication after generation); it passes verification again (novelty search, four-axis scoring, and citation-faithfulness check); and it improves on the grade or score. This re-verification is identical to Stage 7--9.

We compared the pipeline with the regeneration loop off and on, on the same three topics (multimodal, RAG hallucination, and inference efficiency). In every topic, four hypotheses were selected for regeneration, which is the maximum per topic (we do not know how many would have been selected without this maximum): 12 of the 34 hypotheses (35.3\%) were selected in total, and 1 of these 12 (8.3\%) was replaced by the regenerated hypothesis. This hypothesis had been selected because its citation-faithfulness grade was ``none,'' and the replacement had a higher grade. This one replaced hypothesis did not appear in a later paper. We kept the loop off by default, based on the low replacement rate and non-appearance.

\paragraph{Self-assessment has no predictive power.} The critic panel gives each hypothesis an overall quality score (Stage 8 of Section~\ref{sec:pipeline}; the mean of five scores, defined in Section~\ref{sec:mainresults}). We checked whether this score is higher for the hypotheses that later appeared (matched a paper published after the freeze date) than for those that did not (Section~\ref{sec:mainresults}). For each of the nine topics, we compared the mean score of the two groups. In some topics the hypotheses that appeared have the higher mean, while in others those that did not appear have the higher mean (the per-topic difference between the two means ranges from $-0.175$ to $+0.377$). Because the direction differs by topic, the score does not tell whether a hypothesis will appear. We also checked whether the differences could be random variation. With nine topics, we could not tell random variation from a real pattern (Section~\ref{sec:mainresults}).

\section{Discussion and Limitations}
\label{sec:discussion}

\paragraph{What we confirmed and what we could not.} Of the 103 hypotheses ShadowMiner produced, 40 (39\%) appeared in papers published after the freeze date (later papers; Section~\ref{sec:mainresults}). The generation model cannot have learned these papers, because they were published after the freeze date. This design does not rule out other sources of contamination (see ``Contamination boundary'' below). The critic panel labeled each hypothesis ``exists'' or ``new'' after seeing the similar papers found by the novelty search (Stage 8 of Section~\ref{sec:pipeline}). The hypotheses labeled ``exists'' appeared in later papers more often than those labeled ``new'' (Section~\ref{sec:mainresults}). When \texttt{o3}, which is not a member of the critic panel, labeled the hypotheses again, the direction was the same (Section~\ref{sec:reliability}). The higher appearance rate of the hypotheses labeled ``exists'' was not statistically significant, whether with the labels of the critic panel ($p=0.054$) or with the labels of \texttt{o3} ($p=0.130$). This observation therefore has a consistent direction, but we cannot rule out that it reflects random variation rather than a true effect. The critic panel also assigns an overall quality score to each hypothesis. For each topic, we compared the mean overall quality score of the hypotheses that appeared in later papers with that of the hypotheses that did not. In some topics the former had the higher mean, and in others the latter did (Sections~\ref{sec:mainresults} and~\ref{sec:negative}). The limitations of these results follow.

\paragraph{Contamination boundary.} In this study, contamination is handled only by the freeze date (Section~\ref{sec:setup}). The freeze date has four limitations. (1) It cannot block contamination at the level of ideas: the generation model may have absorbed the idea of a paper without the paper itself, and a paper published after the freeze date may have an earlier preprint that was already in the training data. (2) A study of the date-cutoff method [37] points out that the time when the training data was collected and the time when the model was trained can differ, so that information from after the cutoff can still reach the model indirectly. (3) We set the freeze date from the release date of the generation model (\texttt{glm-4.7}, 2025-12-22), because we could not determine its training cutoff from public sources. A model is released after training and refinement, so the release date is later than the training cutoff; using it as the freeze date is therefore a conservative choice for contamination purposes. As a result, the answers are limited to papers published in the period of about nine months between the freeze date (2026-01-01) and the time of the evaluation. (4) The publication date used for the freeze date has different meanings in the four sources of the corpus. In arxiv, it is the date of the first public release, which matches the purpose of the freeze date. In openalex and acl, it is the date of publication in a journal or in proceedings, which can be later than the first public release. We therefore report, alongside the actual appearance rate of 0.388 (40/103), the rate computed with arxiv alone, 0.262 (27/103), which is a lower bound (Section~\ref{sec:mainresults}).

\paragraph{A data defect we found and corrected.} While reviewing the matched papers behind the actual appearance rate, we found that one matched paper had been indexed twice, with the second copy carrying its re-crawl date instead of its true publication date; this let it pass the freeze-date filter despite predating the freeze date. We excluded this match; the corrected rate is 0.379 (39/103), 0.9 percentage points lower than the uncorrected 0.388. We checked as far as our stored data allowed for the same defect elsewhere --- an exhaustive check of the main experiment's 40 matches (1 case) and of the candidate pools used in the negative control of Section~\ref{sec:reliability} (0 cases) --- but the results of Section~\ref{sec:negative} and the A1 ablation do not log individual match dates, so we could not check them the same way.

\paragraph{Limitations of the measurement tool.} We have not yet verified the judgments by human scoring, so the absolute accuracy of the judging tool (the LLM judge) is unknown (Section~\ref{sec:reliability}). What we confirmed in Section~\ref{sec:reliability} is only internal consistency (judgment agreement), the tendency toward false positives in the negative control, and agreement between models; this must not be read as having verified the accuracy of the judgments. Moreover, of the three members of the critic panel, \texttt{glm-5.1} is a model released in 2026-04 and may already have been trained on papers up to early 2026. If so, the model may use its knowledge of papers published after the freeze date (2026-01-01) when it gives its judgments. For the overall quality score, the model might give a high score to a hypothesis that matches a paper it knows (a paper published after the freeze date). For the novelty labels, the model might label such a hypothesis ``exists,'' because it knows a paper that matches it. This bias would make the hypotheses labeled ``exists'' appear in later papers more often, which is what we observed (Section~\ref{sec:mainresults}). We therefore cannot rule out that this bias caused the observation. \texttt{o3}, which relabeled the hypotheses, has a training cutoff of 2024-06-01 [49], which is before the freeze date (2026-01-01), so it cannot know papers published after the freeze date. With the labels of \texttt{o3}, the hypotheses labeled ``exists'' still appeared in later papers more often, but the difference was not statistically significant (Section~\ref{sec:reliability}). We could not determine from public sources the training cutoffs of the other two members of the critic panel (\texttt{gpt-5-mini} and \texttt{glm-4.6}), so the same risk cannot be ruled out for them.

\paragraph{Not appearing is not necessarily being ``wrong.''} The actual-appearance judgment retrieves at most six candidate papers and judges only the top five (Section~\ref{sec:mainresults}). A paper that matches the hypothesis may not be among these five. In that case \texttt{o3} never sees the paper, and the hypothesis is recorded as not having appeared. The 63 of the 103 hypotheses that did not appear (Section~\ref{sec:mainresults}) thus consist of two groups: hypotheses that no paper published after the freeze date matches, and hypotheses whose matching paper the search missed. The number of non-appearing hypotheses is therefore not a count of prediction failures.

\paragraph{Scope and operational limits.} ShadowMiner was tested in a single AI/CS domain only. One of the LLM providers used for judging and generation (the GLM family) has defects --- it returns blank responses on certain sensitive topics, and characters of another language appear in its output. The critic panel mixes several providers, and a judge that fails is not replaced by another model (Table A1, item 4). In the actual-appearance experiment, for 11 of the 103 hypotheses, only two of the three judges returned a result, so the novelty label and quality score came from just those two (all in the \texttt{agents} topic); Appendix~\ref{app:robustness} shows that this does not change the direction of the results of Sections~\ref{sec:mainresults} and~\ref{sec:negative}. Unlike the critic panel, other stages --- such as knowledge-graph extraction and problem-and-hypothesis generation --- call only a single model. If GLM fails, the same request is sent to OpenAI (\texttt{gpt-5-mini}) as a fallback. In this experiment, the fallback was used once, on the \texttt{multimodal} topic, when GLM returned a blank response. We could not determine the effect of this one case on the results. Each hypothesis has citations attached, but the cited documents may not actually support its claim. In Section~\ref{sec:reliability}, we checked the reliability of the other two LLM judgments (the actual-appearance judgment and the novelty labels), but we did not investigate the reliability of this citation-faithfulness check.

\newpage
\section*{References}
\footnotesize

[1] Zhao, K., Kong, L., Xu, F., \& Li, Y. (2026). \emph{Agentic-Ideation: Sample Efficient Agentic Trajectories Synthesis for Scientific Ideation Agents}. arXiv:2606.31229.

[2] Gottweis, J., Weng, W.-H., Daryin, A., et al. (2026). \emph{Accelerating Scientific Discovery with Co-Scientist}. Nature (2026). arXiv:2502.18864.

[3] Lu, C., Lu, C., Lange, R. T., Foerster, J., Clune, J., \& Ha, D. (2024). \emph{The AI Scientist: Towards Fully Automated Open-Ended Scientific Discovery}. Sakana AI. arXiv:2408.06292.

[4] Ibrahim, F. N., Azad, N. S., \& Baten, R. A. (2026). \emph{Anchorless Diversification for Parallel LLM Ideation}. arXiv:2605.30150.

[5] Li, Y., Shao, C., Liu, X., et al. (2026). \emph{AutoSOTA: An End-to-End Automated Research System for State-of-the-Art AI Model Discovery}. arXiv:2604.05550.

[6] Deng, Y., Brucks, M., \& Toubia, O. (2026). \emph{Examining and Addressing Barriers to Diversity in LLM-Generated Ideas}. arXiv:2602.20408.

[7] Li, L., Xu, W., Guo, J., et al. (2024). \emph{Chain of Ideas: Revolutionizing Research via Novel Idea Development with LLM Agents}. arXiv:2410.13185.

[8] Lin, L., Pan, X., Wang, R., Zhou, C., Wang, J., Wang, Y., \& Rasovska, I. (2026). \emph{DN-Hypo-Pipeline: An AI-Driven Workflow for Hypothesis Generation via Large Language Models and Scientific Explanations}. arXiv:2606.08532.

[9] Si, C., Yang, Z., Choi, Y., Cand\`es, E., Yang, D., \& Hashimoto, T. (2026). \emph{Towards Execution-Grounded Automated AI Research}. arXiv:2601.14525.

[10] Wang, H., Shojaee, P., Meidani, K., et al. (2026). \emph{Towards Diverse Scientific Hypothesis Search with Large Language Models}. arXiv:2606.10587.

[11] Li, X., Tu, H., Li, X., Zhao, K., Han, X., \& Liu, Z. (2026). \emph{EvoGens: A Population-Based Heuristic Search Framework for Scientific Idea Generation}. arXiv:2605.30961.

[12] Le, X., Qian, P., Zhai, Y., Zhang, X., Liu, Q., Dawei, F., \& Ding, B. (2026). \emph{EvoNarrator: Modeling Scientific Evolution for Feasible Hypothesis Generation}. ACL 2026.

[13] Xiong, X., Ren, Y., \& Xiong, D. (2026). \emph{EvoSci: A Bio-Inspired Multi-Agent Framework for the Evolution of Scientific Discovery}. arXiv:2605.24018.

[14] Falck, D., Sabri, S., Surina, A., et al. (2026). \emph{Training AI Scientists to Replicate Research}. arXiv:2608.13331.

[15] Gao, S., Xia, Y., Liu, S., \& Xiong, H. (2026). \emph{Graphs of Research: Citation Evolution Graphs as Supervision for Research Idea Generation}. arXiv:2605.14790.

[16] Pal, S., Sourav, S., Ghosal, T., \& Buehler, M. J. (2026). \emph{Graph-Native Reinforcement Learning Enables Traceable Scientific Hypothesis Generation through Conceptual Recombination}. arXiv:2607.00924.

[17] Jiang, B. (2026). \emph{HindSight: Evaluating LLM-Generated Research Ideas via Future Impact}. arXiv:2603.15164.

[18] Zhu, Z., Xue, Y., Chen, X., Zhou, D., Tang, J., Schuurmans, D., \& Dai, H. (2023). \emph{Large Language Models can Learn Rules}. arXiv:2310.07064.

[19] Qian, Z., Lei, J., Wang, Y., \& Cao, N. (2026). \emph{HypoForge: A Self-Improving Multi-Agent Framework for Automated Hypothesis Generation and Testing via Scientific Skill Learning}. arXiv:2608.25770.

[20] Zhou, Y., Liu, H., Srivastava, T., Mei, H., \& Tan, C. (2024). \emph{Hypothesis Generation with Large Language Models}. arXiv:2404.04326.

[21] Gumma, V., Majumder, N., Sinhahajari, S., \& Poria, S. (2026). \emph{IDEAgent: Agentic Quality-Diversity Search for Research Idea Generation}. arXiv:2607.22375.

[22] Zhou, Y., Yang, Q., Li, Y., et al. (2026). \emph{Ideas Have Genomes: Benchmarking Scientific Lineage Reasoning and Lineage-Grounded Idea Generation}. arXiv:2607.08758.

[23] Qiao, S., Wei, Y., Wang, X., et al. (2026). \emph{InnoEval: On Research Idea Evaluation as a Knowledge-Grounded, Multi-Perspective Reasoning Problem}. ICML 2026. arXiv:2602.14367.

[24] Xiong, G., Xie, E., Shariatmadari, A. H., Guo, S., Bekiranov, S., \& Zhang, A. (2024). \emph{Improving Scientific Hypothesis Generation with Knowledge Grounded Large Language Models}. arXiv:2411.02382.

[25] Wang, Z., Yuan, A., Yao, Y., et al. (2026). \emph{What Proves You Wrong: Benchmarking Language Models on Falsifiable Research Ideation}. arXiv:2608.22948.

[26] Yang, Z., Liu, W., Gao, B., et al. (2024). \emph{MOOSE-Chem: Large Language Models for Rediscovering Unseen Chemistry Scientific Hypotheses}. arXiv:2410.07076.

[27] Oh, J., Kim, B., Li, J., Park, Y. J., \& Park, J.-S. (2026). \emph{Multi-Persona Debate System for Automated Scientific Hypothesis Generation}. arXiv:2605.23917.

[28] Hu, X., Fu, H., Wang, J., et al. (2024). \emph{Nova: An Iterative Planning and Search Approach to Enhance Novelty and Diversity of LLM Generated Ideas}. arXiv:2410.14255.

[29] Schopf, T., \& F\"arber, M. (2026). \emph{Is this Idea Novel? An Automated Benchmark for Judgment of Research Ideas}. LREC 2026. arXiv:2603.10303.

[30] Baek, J., Jauhar, S. K., Cucerzan, S., \& Hwang, S. J. (2025). \emph{ResearchAgent: Iterative Research Idea Generation over Scientific Literature with Large Language Models}. NAACL 2025. arXiv:2404.07738.

[31] Zhao, Z., Wen, H., Ma, J., et al. (2026). \emph{ResearchEVO: An End-to-End Framework for Automated Scientific Discovery and Documentation}. arXiv:2604.05587.

[32] Ghafarollahi, A., \& Buehler, M. J. (2024). \emph{SciAgents: Automating Scientific Discovery through Multi-Agent Intelligent Graph Reasoning}. arXiv:2409.05556.

[33] Xu, Y., Zhang, K., Yu, Y., et al. (2026). \emph{Scaling Scientific Discovery Environments for Turn-Level Agentic RL}. arXiv:2607.28990.

[34] Radensky, M., Shahid, S., Fok, R., Siangliulue, P., Hope, T., \& Weld, D. S. (2024). \emph{Scideator: Human-LLM Scientific Idea Generation Grounded in Research-Paper Facet Recombination}. arXiv:2409.14634.

[35] Wang, Q., Downey, D., Ji, H., \& Hope, T. (2024). \emph{SciMON: Scientific Inspiration Machines Optimized for Novelty}. ACL 2024. arXiv:2305.14259.

[36] Swanson, D. R. (1986). \emph{Fish oil, Raynaud's syndrome, and undiscovered public knowledge}. Perspectives in Biology and Medicine, 30(1), 7--18.

[37] Cheng, J., Marone, M., Weller, O., Lawrie, D., Khashabi, D., \& Van Durme, B. (2024). \emph{Dated Data: Tracing Knowledge Cutoffs in Large Language Models}. COLM 2024. arXiv:2403.12958.

[38] Liu, H., Tjiaranata, F. A., \& Tan, C. (2026). \emph{VERITAS: Towards a General-Purpose Replication Tool for Scientific Research}. arXiv:2607.02931.

[39] Ding, T., Nannapaneni, A., Liu, B., \& Zhang, L. (2026). \emph{Autonomous Research Agents: A Survey of AI Scientists and the Verification Gap}. arXiv:2608.05179.

[40] Zwicky, F. (1969). \emph{Discovery, Invention, Research Through the Morphological Approach}. Macmillan.

[41] Hou, Y., Jochim, C., Gleize, M., Bonin, F., \& Ganguly, D. (2021). \emph{TDMSci: A Specialized Corpus for Scientific Literature Entity Tagging of Tasks, Datasets, and Metrics}. EACL 2021. arXiv:2101.10273.

[42] Jain, S., van Zuylen, M., Hajishirzi, H., \& Beltagy, I. (2020). \emph{SciREX: A Challenge Dataset for Document-Level Information Extraction}. ACL 2020. arXiv:2005.00512.

[43] Luan, Y., He, L., Ostendorf, M., \& Hajishirzi, H. (2018). \emph{Multi-Task Identification of Entities, Relations, and Coreference for Scientific Knowledge Graph Construction}. EMNLP 2018. arXiv:1808.09602.

[44] Zhang, Q., Chen, Z., Pan, H., Caragea, C., Latecki, L. J., \& Dragut, E. (2024). \emph{SciER: An Entity and Relation Extraction Dataset for Datasets, Methods, and Tasks in Scientific Documents}. EMNLP 2024. arXiv:2410.21155.

[45] Lin, Z., Yin, Y., Liu, L., \& Wang, D. (2023). \emph{SciSciNet: A Large-Scale Open Data Lake for the Science of Science Research}. Scientific Data, 10, 315.

[46] Borrego, A., Dess\`i, D., Ayala, D., Hern\'andez, I., Osborne, F., Reforgiato Recupero, D., Buscaldi, D., Ruiz, D., \& Motta, E. (2025). \emph{Research Hypothesis Generation over Scientific Knowledge Graphs}. Knowledge-Based Systems, 315, 113280.

[47] Wu, L., Wang, D., \& Evans, J. A. (2019). \emph{Large Teams Develop and Small Teams Disrupt Science and Technology}. Nature, 566(7744), 378--382.

[48] Qiu, Y., Zhang, H., Xu, Z., Li, M., Song, D., Wang, Z., \& Zhang, K. (2025). \emph{AI Idea Bench 2025: AI Research Idea Generation Benchmark}. arXiv:2504.14191.

[49] OpenAI. (2025). \emph{o3 Model} (snapshot o3-2025-04-16; knowledge cutoff June 1, 2024). OpenAI API documentation. \url{https://developers.openai.com/api/docs/models/o3} (accessed 2026-09-19).
\normalsize

\appendix

\section{Techniques and Their Sources}
\label{app:techniques}

This is the full table of the techniques summarized in Section~\ref{sec:techniques}. In this table, DGG means the described graph gap and DGG-SP the described graph gap with source papers (Sections~\ref{sec:pipeline} and~\ref{sec:negative}).

\begin{landscape}
\begin{footnotesize}
\noindent\textbf{Table A1.} List of techniques and their sources. For each technique: its source (what was taken from where), its implementation, its measured results, and whether it was finally adopted.\par\vspace{4pt}

\begin{longtable}{p{0.3cm}p{2.6cm}p{2.6cm}p{3.2cm}p{3.2cm}p{3.6cm}p{2.4cm}}
\toprule
\# & Technique & What was taken & Source & Implementation & Measured result & Adoption \\
\midrule
\endfirsthead
\toprule
\# & Technique & What was taken & Source & Implementation & Measured result & Adoption \\
\midrule
\endhead
\bottomrule
\endfoot

--- & Graph-gap detection and description --- matrix gap, task transfer, ABC transfer, unresolved limitations & Search for graph gaps by rules rather than by an LLM & Swanson's ABC linking (literature-based discovery [36]) and Zwicky's morphological analysis [40]; the idea of an explicit gap-analysis step has the same aim as the \texttt{Analyse\_Gap} tool of Agentic-Ideation [1] & Four rule-based patterns, no LLM, computation of support, with a minimum support for adoption (2 for the matrix gap, ABC transfer, and unresolved limitations; 3 for task transfer); the result is inserted into the Stage 6 prompt as described graph gaps (DGGs, for reference) & Baseline (with graph-gap hints) 13/36 = 0.361 vs. ablated (hints removed) 13/32 = 0.406, a difference of +4.5pp, not statistically significant ($p=0.702$); the direction differed by topic (one topic dropped sharply, one rose sharply, one unchanged) & Kept as a default component of ShadowMiner \\

1 & Experiment-execution (code-execution verification) loop & A closed loop that runs experiments for a hypothesis and verifies the result by measurement & AutoSOTA [5], VERITAS [38], SciDisco [33], HypoForge [19], and others & Not started & --- & The experiment-execution loop is not considered; only hypothesis generation and evaluation are addressed \\

2 & Critic feedback $\rightarrow$ regeneration loop & For a hypothesis below the threshold, feed back per-axis comments and regenerate once & IDEAgent [21] (repair success rate 93\%, refine 82\%), ResearchAgent [30] (converges after three regenerations) & Citation-faithfulness grade as the first-priority criterion and critic axes as the secondary criterion; one regeneration per hypothesis and a per-topic budget cap; hard-duplicate check; replace the original only on improvement & Test on three topics, 34 hypotheses: targeted 35.3\% of cases but replaced only 8.3\% (1 of 12); the one replacement did not appear in a later paper & Implemented (default off) --- \textbf{not adopted} \\

3 & Parallel multi-hypothesis generation + diversity & Turn one call into N parallel calls; add merging, deduplication, and a diversity metric & Semantic diversity $\times$2.6 with ordinary personas [6]; a one-line instruction to ``stand out from other responses that might be generated for this same task'' raised semantic diversity significantly at about $\times$1.1 token cost [4]; the Yield metric (the size of the largest set of hypotheses that pass the quality threshold and are sufficiently different from one another) is from IDEAgent [21]; generation-by-generation evolving candidate pools are from EvoGens [11] & \texttt{gen\_mode} option (a single call; parallel makes N calls independently at the same time; sequential makes N calls one after another, each told the earlier results and instructed to differ) + merging and deduplication + Yield metric & Single call: groundedness 3.00, appearance rate 0.278 (5/18). Parallel (adopted): groundedness 3.03, appearance rate 0.387 (12/31) --- more hypotheses in total (31 vs. 18) at about the same groundedness. Sequential: groundedness 3.05, appearance rate 0.208 (5/24), waste rate higher than parallel in every topic, 1.0--1.8$\times$ slower & \textbf{Parallel adopted} (made the default; \texttt{gen\_n=4}, duplicate threshold 0.75), sequential dropped \\

4 & Critic multi-provider ensemble & Move away from single-provider self-scoring and score with different providers/models & Prior cases showing the need for ensemble scoring: novelty-judgment benchmark [29], HindSight [17], InnoEval [23], The AI Scientist [3] (five-reviewer ensemble) & Extend \texttt{critic\_models} to a list of \texttt{"provider:model"}, automatically skip unavailable providers, and turn the fallback off for judge calls (so a judge is not replaced by another model on failure) & Switched the operating panel to \texttt{[glm:glm-5.1, openai:gpt-5-mini, glm:glm-4.6]}, removing scoring by the generation model (\texttt{glm-4.7}) itself (two of the three panel members are of the same GLM family, one is OpenAI) & Code completed, then \textbf{actually switched on and operated with this setting} \\

5 & Explicit falsification observation (six-field contract) & Specify in structured form the conditions under which a hypothesis can be falsified & Lit2Test [25] & Design only, not started & --- & Postponed \\

6 & Learning-based capability accumulation (SFT, RL, skill accumulation) & Build up capability through learning rather than a prompt chain & HypoForge [19] (skill accumulation without fine-tuning), GoR [15] and Agentic-Ideation [1] (SFT), Execution-Grounded [9] (RL with execution reward alone leads to mode collapse --- 119 of 128 collapse onto 2), Faraday [14] (RL succeeds only under narrow conditions, such as a five-dimension rubric reward) & Not started & --- & Intentionally omitted --- SFT data is hard to obtain in a general domain, and the risk of mode collapse in RL has been demonstrated \\

7 & Idea genealogy and cross-session duplicate memory & Track generation history to suppress duplicates & GoR [15] and EvoNarrator [12] (generation genealogy), IG-Bench [22] (genealogy evaluation), IDEAgent [21] (five-tuple signature archive) & Not started & --- & Deferred --- cross-session global deduplication matters only when hypotheses are generated continuously as one stream, which does not apply to the setup of this report (an independent run per topic) \\

8 & In-loop oracle (strengthening the external verification loop that intervenes during generation) & A gate that brings external verification into the middle of generation & Survey [39] notes that none of the LLM-era systems has an externally verified in-loop oracle & Not started & --- & Deferred --- the Stage 9 citation-faithfulness check is an LLM check attached after generation, not an external verification that intervenes during generation \\

9 & Giving similar-paper abstracts to the novelty judgment & Extend the information used in the judgment from titles alone to abstracts & (Not an external paper --- an internal fix that makes use of information that had already been obtained but went unused) & Include the abstract (up to 1,000 characters) in the return of \texttt{novelty\_search()}, and add to the critic prompt the instruction to ``judge by the abstract, not the title'' & Zero additional LLM calls; confirmed against the actual search engine (Elasticsearch) that abstracts are really filled in & \textbf{Applied} \\

10 & Attaching sources to graph gaps (DGG $\rightarrow$ DGG-SP) & Turn the one-line DGG into the DGG-SP by adding source-paper numbers & Agentic-Ideation [1] --- an explicit gap-analysis step (\texttt{Analyse\_Gap}) raised the novelty score from 5.86 to 6.74 (only the idea drawn from this result was borrowed) & \texttt{gen\_lead\_format} setting (the actual name in the code); Stage 5 returns a record (dict) containing the source-paper numbers and Stage 6 displays it & Comparison on three topics: groundedness about the same (DGG 3.24 against DGG-SP 3.22), actual appearance rate higher with the DGG-SP but not statistically significant (0.412 against 0.472, $p=0.611$); cost increased in every topic; no benefit large enough to justify the added cost & \textbf{Not adopted}; default \texttt{flat} (DGG) kept (the code is preserved behind a setting) \\

11 & Cut priority for the entity list in Stage 6 & Replace the cut criterion from alphabetical order with a three-tier selection: recency (entities that appear in triples read in this run), consistency (entities named by DGGs and the limitation summary, so that entities referenced by other prompt blocks are not dropped from the list), and triple count & (Not an external paper --- an internal fix that removes a bias that had no research basis) & Replace the selection logic of \texttt{render\_kg\_for\_prompt()} (cap 40 $\rightarrow$ 60, hard cap 180) & Ran the function from the operating code directly on 43 stored real KGs: alphabetical cutting was removed in all 82 types that exceeded the cap, and the consistency criterion rescued a total of 112 concepts from being cut across 23 topic runs & \textbf{Applied} \\

12 & Extending the novelty comparison from abstracts to full text & Stage 6 hypotheses are built from triples extracted from the first 12,000 characters of each paper's full text, but the novelty comparison of Stages 7 and 8 sees only the abstracts of existing papers --- remove this asymmetry & (Not an external paper --- an internal improvement proposal) & Not started (not free of cost: the similar papers found by novelty search can number up to 36 per topic, while the full-text parsing tool's processing cap is 5) & --- & To be reconsidered later (revisit when GPU resources are strengthened) \\

\end{longtable}
\end{footnotesize}
\end{landscape}

\section{Robustness Check: Hypotheses Judged by Two Judges}
\label{app:robustness}

\paragraph{Hypotheses judged by two judges.} The critic panel assigns a novelty label to each hypothesis using three judges, but for an unknown reason, 11 of the 103 hypotheses, all in the \texttt{agents} topic, were judged by only two. The novelty labels of these 11 hypotheses are ``new'' for 8 and ``exists'' for 3. We recomputed the results of Section~\ref{sec:mainresults} without these 11 hypotheses. The appearance rate is the number of hypotheses with a given label that appeared in papers published after the freeze date, divided by the number of hypotheses with that label, and the significance is from a two-proportion $z$-test. Table B1 shows this comparison for four cases. In all four, hypotheses labeled ``exists'' appeared in papers published after the freeze date more often than those labeled ``new.''

\begin{table}[H]
\centering
\textbf{Table B1.} Exists/new appearance-rate difference across the four cases of Appendix~\ref{app:robustness}.
\vspace{4pt}

\footnotesize
\begin{tabular}{L{5.2cm}C{1.7cm}C{1.7cm}L{1.3cm}L{2.0cm}}
\toprule
Hypotheses included & Exists appearance rate & New appearance rate & Difference & Significance \\
\midrule
(A) All hypotheses, with no exclusions (100) & 0.538 (14/26) & 0.338 (25/74) & +0.201 & $z=1.80$, $p=0.071$ \\
(B) Of (A), excluding the two-judge hypotheses (89) & 0.522 (12/23) & 0.318 (21/66) & +0.204 & $z=1.74$, $p=0.082$ \\
(C) Of (A), excluding those affected by the freeze-date leak discussed in Section~\ref{sec:mainresults} (92) & 0.571 (12/21) & 0.338 (24/71) & +0.233 & $z=1.93$, $p=0.054$ \\
(D) Of (C), excluding the two-judge hypotheses (82) & 0.579 (11/19) & 0.317 (20/63) & +0.261 & $z=2.06$, $p=0.039$ \\
\bottomrule
\end{tabular}
\end{table}

For the overall quality score, we repeated the Section~\ref{sec:mainresults} comparison: for each topic, the mean score of the hypotheses that appeared (matched a paper published after the freeze date) versus the mean score of those that did not. This time we used only the hypotheses judged by three judges. This excludes the \texttt{agents} topic, since all 11 of its hypotheses were judged by two judges; the comparison covers the other eight topics. As in Section~\ref{sec:mainresults}, the hypotheses that appeared in later papers had the higher mean score in some topics, and those that did not had the higher mean in others; the per-topic difference between the two means ranges from $-0.175$ to $+0.377$, the same range as in Section~\ref{sec:mainresults}. We also repeated this check (Section~\ref{sec:mainresults}): the correlation is $r=0.191$ ($p=0.650$) here, compared with $r=0.049$ ($p=0.901$) in Section~\ref{sec:mainresults}. Because this correlation is weak and not statistically significant, we still cannot tell whether the per-topic differences reflect random variation or a real pattern. The conclusions of Section~\ref{sec:mainresults} do not change.

\section{Reproducibility}
\label{app:repro}

\paragraph{Time and completion rate.} The median time per topic was 91 minutes (range 65--118 minutes), and all nine topics were completed (9/9) with zero failures (Section~\ref{sec:setup}).

\paragraph{Hyperparameters.}

\setcounter{table}{6}
\begin{table}[H]
\caption{Settings applied to the actual-appearance experiment of Section~\ref{sec:mainresults} (nine topics).}
\label{tab:table7}
\centering
\footnotesize
\begin{tabular}{L{5.0cm}L{7.0cm}}
\toprule
Item & Value \\
\midrule
Hypothesis-generation model & \texttt{glm-4.7} (Stage 6) \\
Hypothesis-generation mode & parallel (\texttt{gen\_n=4}; duplicate threshold: embedding cosine similarity $\geq$ 0.75) (Stage 6) \\
Critic judging panel & \texttt{glm-5.1} / \texttt{gpt-5-mini} / \texttt{glm-4.6} (three judges, mixed providers) (Stage 8) \\
Citation-faithfulness and factuality check & On (Stage 9) \\
Actual-appearance evaluation judge model & \texttt{o3} (three-vote majority) \\
Narrow graph / wide graph & 1-hop / 2-hop (Stage 3) \\
Document-importance weight cap & 0.15 relative to relevance (Stage 1) \\
Entity-list cap in the hypothesis-generation prompt & 60 per type (hard cap 180; Stage 6; Table A1, item 11) \\
Novelty search top K & 3 (Stage 7) \\
Actual-appearance evaluation search candidates / checked & up to 6 papers / top 5 \\
Actual-appearance evaluation search query & The prediction and the verification method of each hypothesis \\
Contamination boundary (freeze date) & 2026-01-01 (based on the generation model's release date) \\
Input documents per topic & up to 40 (knowledge-graph extraction on the top 30; Stage 2) \\
Problems / hypotheses per topic & up to 6 / up to 12 (Stage 6) \\
\bottomrule
\end{tabular}
\end{table}

Table~\ref{tab:table7} gives the settings needed for reproduction.

\paragraph{Cost.} The judging cost of the actual-appearance evaluation was \$0.52--0.70 per topic.

\paragraph{Availability.} The code and data are not released.

\paragraph{Writing assistance.} Parts of this report's text were drafted and revised with the assistance of Claude Code (Anthropic), used as a writing and editing aid. The authors are responsible for all content.

\end{document}